\documentclass{article}
\usepackage{spconf, amsmath, graphicx, amssymb, mathtools, pbox, tabu, enumitem, float, placeins}
\usepackage[skip=2pt, font=small]{caption}

\title{LEARNING TO SEGMENT ON TINY DATASETS: A NEW SHAPE MODEL}

\name{Maxime Tremblay, Andr\'{e} Zaccarin}
\address{Computer Vision and Systems Laboratory, Universit\'{e} Laval\\
 Quebec City (QC), Canada}

\newcommand{\Dsty}{\mathcal{D}} 
 
\newcommand{\Msty}{\mathcal{M}} 

\renewcommand{\deg}{^\circ}

\begin{document}

\maketitle

\begin{abstract}

Current object segmentation algorithms are based on the hypothesis that one has access to a very large amount of data. In this paper, we aim to segment objects using only tiny datasets. To this extent, we propose a new automatic part-based object segmentation algorithm for non-deformable and semi-deformable objects in natural backgrounds. We have developed a novel shape descriptor which models the local boundaries of an object's part. This shape descriptor is used in a bag-of-words approach for object detection. Once the detection process is performed, we use the background and foreground likelihood given by our trained shape model, and the information from the image content, to define a dense CRF model. We use a mean field approximation to solve it and thus segment the object of interest. Performance evaluated on different datasets shows that our approach can sometimes achieve results near state-of-the-art techniques based on big data while requiring only a tiny training set.

\end{abstract}

\begin{keywords}
object segmentation, shape model, tiny data set, bag-of-words, dense CRF
\end{keywords}
%

\section{Introduction}

Object segmentation algorithms based on deep neural networks (such as DeepMask~\cite{Pinheiro2015} and SharpMask~\cite{Pinheiro2016}) are extremely powerful and greatly outperform former state-of-the-art approaches. Deep learning requires massive datasets to train a system and, even when it is pre-trained on another dataset, the network needs to be fine-tuned with at least a medium-size dataset to perform as advertised. In this work, we focused on learning on very small datasets, which we call \emph{tiny datasets}. More specifically, our tiny datasets contain fewer than 50 training samples. Given this constraint, an interesting approach is to model an object as a set of parts~\cite{Leibe2008, Marszaek2012}. The shape of the object is implicitly or, sometimes, explicitly preserved through detected features of interest. These methods are intrinsically designed for detection and segmentation of occluded objects. Furthermore, they generally do not need a large number of images for training. However, the final segmentation will be, at best, a weighted sum of deformed templates, which means the labeling accuracy near the boundary regions will be fairly low. Inspired by this, we focus our work on a bag-of-words method for which the segmentation problem is solved globally using local pixel-level information.

We propose an approach inspired by the Implicit Shape Model~\cite{Leibe2008} in which an object is represented by a clustering of appearance features, their relative position to the training objects, and ground truth segmentation masks. However, we believed that a shape model should be used to identify which regions are part of the object and which regions are outside the object, and that object boundaries should be enforced by the image content. Accordingly, while learning the object appearance and shape models, we preserve the shape information into a low-level descriptor. This low-level descriptor, which we call a \emph{shape descriptor}, captures the shape of the object boundaries. Once an object is detected, the shape descriptors provide pixel-wise foreground and background likelihood. These are used to segment the objects with a dense CRF model which we solve using a mean field approximation~\cite{Krahenbuhl2011}. We will refer to our approach as Boundary Shape Model (BSM).

The main contributions of this paper are the introduction of a shape descriptor which models strong boundaries, and an object segmentation framework which is trainable with a tiny set of samples.

\section{Shape Descriptor}
\label{sec:shape_descriptor}

We capture the shape of an object part with a descriptor that we designed to take into account the location and orientation of its strong boundaries.
We propose to model shapes with a quantized SIFT descriptor on the ground truth binary masks of the objects. As a reminder, a SIFT descriptor~\cite{Lowe2004} is an array of 4x4 cells, each cell being an 8-bins histogram of local gradient orientations of a pixel patch around a detected feature. Since we extract the SIFT descriptor on a binary mask, the gradients do not encapsulate texture information, but shape information. The quantization of this descriptor accentuates local boundaries and captures their direction.

Our shape descriptor is created by first computing a SIFT descriptor on the binary-mask of an object. This SIFT-descriptor is extracted at the same image coordinate of a previously detected appearance-based feature on the object.
Quantization of the SIFT vector parameters generates the shape descriptor $\Dsty_k$ of feature $k$: 
\begin{equation} 
\Dsty_{k}(i) = \frac{\text{sgn}\left( \frac{d_{k}(i)}{m} - 1\right) + 1}{2}, \quad m = \beta \max_i d_{k}(i) 
\label{eq:sd_qz} 
\end{equation} 
where $d_{k}(i)$ is the $i^{th}$ bin of the SIFT descriptor (here $i$ indexes the 128 values or bins of the SIFT descriptor). The threshold value, $m$, is $\beta$ times the descriptor maximum value. We have chosen empirically a value of 0.4 for this $\beta$ factor. 

Each histogram bin of the shape descriptor, $\Dsty_{k}(i)$, is either 0 or 1. A cell inside or outside the object has an empty histogram. A cell with non-zero histogram values describes a boundary. The activated orientation bins (i.e. non-zero elements) of a boundary cell's histogram determine if a neighboring cell with an empty histogram is either inside the object (foreground) or outside the object (background). From $D_k$, we compute the strength $\upsilon_k$ of the foreground or background hypothesis for each of the 16 cells (foreground if $\upsilon_k(i)>0$, background if $\upsilon_k(i)<0$ and 0 otherwise), as 
\small
\begin{equation} 
\upsilon_k(i) = \sum_{j=0}^{7}{h_j(\left(j+4\right) \bmod 8) - h_j(j)} , \quad \text{if} \sum_{k=0}^7 h_i(k) = 0
\label{eq:upsilon_1}
\end{equation}
\normalsize
where $i$ is the index of a cell, $j$ is the index of a relative neighbor of that cell, and $h_i(k)$ is the $k^{th}$ bin of a histogram cell $i$. The histograms are indexed from 0 to 7 clockwise, starting on the right side of the observed cell, and the orientation bins, also indexed from 0 to 7, range from $0\deg$ to $315\deg$ by $45\deg$ increments. Note that a non-empty cell has $\upsilon_k(i) = 0$.

Eq.~\ref{eq:upsilon_1} indicates that to calculate $\upsilon_k(i)$, we only take into account the histogram bins of a neighboring cell that are in the same direction as this neighboring cell with respect to cell $i$. For example, for the cell to the left of cell $i$, which is indexed by the value $j=4$, we take into account the values of the $0\deg$ ($h_{j=4}(0)$) and the $180\deg$ ($h_{j=4}(4)$) bins of that cell.

Finally, it is possible that no adjacent cell $j$ of a cell $i$ contains gradient information. In that case, we propagate values from its neighbors to these cells with
\begin{equation} 
\upsilon_i = \max_j \upsilon_j + \min_j \upsilon_j, \quad \text{iff} \sum_{j=0}^7\sum_{k=0}^7h_j(k) = 0 
\label{eq:upsilon_2}
\end{equation} 
until all $\upsilon_k(i)$ are correctly evaluated. From Eq.~(\ref{eq:upsilon_2}) it is clear that if a foreground or background cell is surrounded by foreground or background cells, the strength of the foreground or background hypothesis should be similar to that of its neighboring cell. This allows for coherent $\upsilon_k(i)$ values for areas far from the contours of the object. 

\begin{figure} 
\centering
	\includegraphics[width=8.5cm]{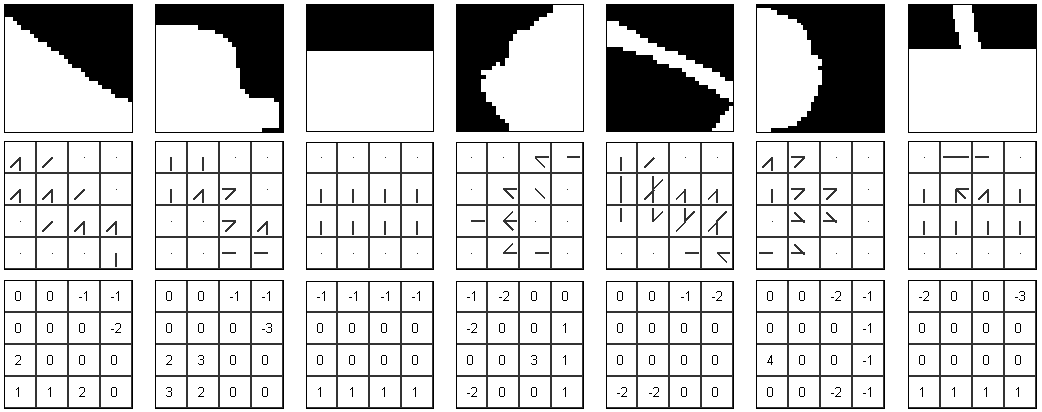}
	\caption{First row: examples of segmentation masks. Second row:  Shape descriptors; each square represents a histogram and each line inside one of these squares represents an activated orientation bin. Third row: $\upsilon$ value for each histogram. Our shape descriptor can represent different shapes while keeping these representations simple. Observe that the $\upsilon(i)$ values give accurate cues about the foreground and background.} 
	\label{fig:sd_examples}
\end{figure}

Fig.~\ref{fig:sd_examples} shows that the shape descriptor can model straight and curved lines while also being robust to small shape variations. Each histogram of the descriptor represents a group of pixels which means that the exact location of the local boundaries are unknown; this confers more smoothness to the representation since a small offset in the boundary position estimation should not hinder the segmentation process. 

\section{Object Model}

We use this new shape descriptor to model the shape of an object class and combine that model with an appearance model in order to detect and segment objects. Our approach is inspired by~\cite{Leibe2008}. We model each object class independently using a bag-of-words approach in which, contrarily to standard bag-of-words approaches (such as~\cite{Fergus2010}), we solely sample from the foreground. We model the appearance of the object by extracting descriptors on detected points of interest in all of the images from the training set. We also extract a descriptor on the binary ground truth image on the location of these features to create shape descriptors (Sec.~\ref{sec:shape_descriptor}). The clustering of the appearance descriptors creates a part-based model of an object. For each cluster, we create a shape model by clustering all the shape descriptors associated with its appearance descriptors.  Using this appearance model, we can detect objects from a specific class in an image by voting. Once an object is detected, we use the information from the shape model of each activated appearance codeword to compute foreground and background likelihood, and then segment the object. 

\subsection{Learning the Models}
\label{sec:learning_model}

The first step of the learning phase is the sampling of all images of an object-class training set. This is carried out with a Harris-Laplace corner detector combined with a SIFT detector, to identify regions of interest (ROIs) on the objects. For each detected feature, the distance from the center of the object, its scale and the size of the object are preserved; this is called an occurrence. SIFT descriptors centered on these features are extracted.

The descriptors are then clustered using the agglomerative clustering algorithm described by~\cite{Leibe2008}. Succinctly, the algorithm begins with each descriptor as its own cluster, then it iteratively merges reciprocate nearest-neighbor pairs of clusters if they are similar enough (using a threshold $t$, where typically $t=0.7$). The similarity between two descriptors is based on Euclidean metric. This clustering creates the appearance \emph{codebook}, a set of \emph{codewords} that represents the object's parts. Each codeword $C_i$ is therefore the average of a set of appearance descriptors.

We also extract SIFT descriptors on the binary mask of the ground truth images at the same coordinate of the detected features. Each appearance descriptor is therefore paired to a SIFT descriptor from a binary image, and, consequently, there is a set of these SIFT descriptors associated with each appearance codeword. We cluster these descriptors and quantize them (following equation~\ref{eq:sd_qz}) to generate a set of shape descriptors for each codeword $C_i$. We denote this shape codebook by $\Dsty_{i(n)}$. In practise, this allows shape variations to be captured for each object's part. 

\subsection{Object detection}
\label{sec:obj_detect}
To detect an object with the trained model, we first sample SIFT descriptors in the image using the same feature detectors as in the learning phase. Then, each of these descriptors is compared with the codewords in the appearance codebook. All matches between a descriptor and a codeword that have a similarity of at least $t$ casts one vote for each of its associated occurrences. Each vote uses the features' distance and scale values preserved at the learning phase. Local maxima in the voting space point to an object approximate size and location. We weight each vote by
\begin{equation} 
\label{eq:weight}
w_{k,i} = \frac{1}{\left| \Msty_{k} \right|} \: \frac{1}{\left| O_{i} \right|}
\end{equation}
where $\left| \Msty_{k} \right|$ is the number of codewords matched with a detected feature $f_k$ and $\left| O_{i} \right|$ the number of occurrences of codeword $C_i$.

Once votes are cast, maxima in the voting space, which are considered good object hypotheses, are retrieved by a mean-shift mode estimation. These hypotheses are then refined using uniform sampling inside a ROI which is determined by the aforementioned maximum in the voting space and by the size of the training objects. This sampling increases the number of matched features and therefore the image area for which the shape model will provide foreground and background likelihood information.
\setlength\tabcolsep{1pt}
\begin{figure*}[t]
\centering
\begin{tabular}{cccccc}
\includegraphics[height=1.8cm]{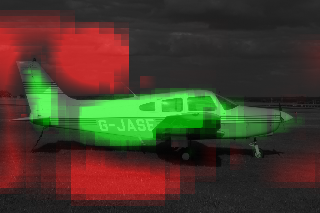}&
\includegraphics[height=1.8cm]{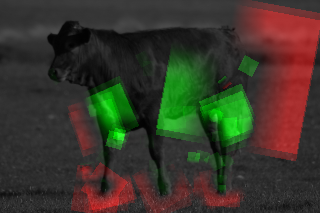}&
\includegraphics[height=1.8cm]{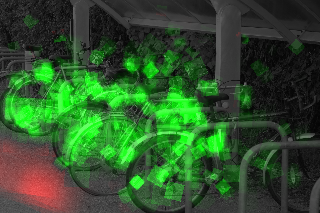}&
\includegraphics[height=1.8cm]{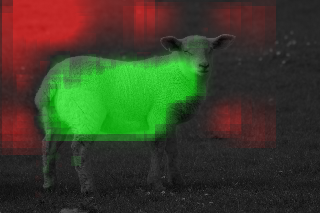}&
\includegraphics[height=1.8cm]{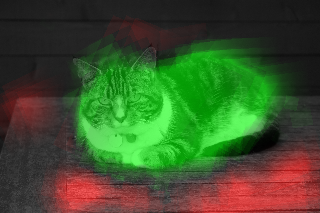}&
\includegraphics[height=1.8cm]{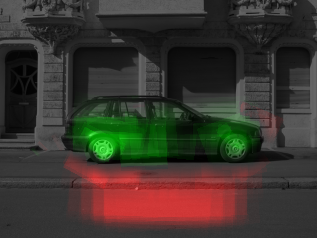}\\
\includegraphics[height=1.8cm]{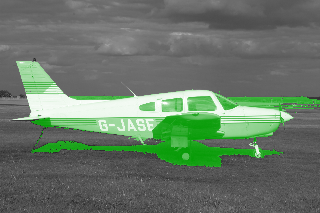}&
\includegraphics[height=1.8cm]{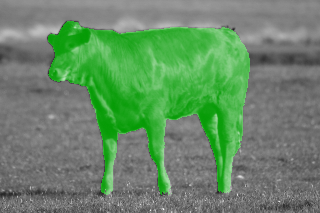}&
\includegraphics[height=1.8cm]{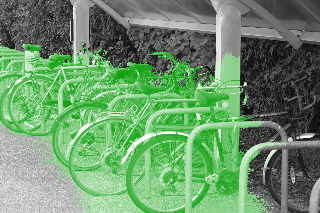}&
\includegraphics[height=1.8cm]{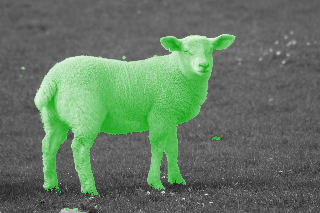}&
\includegraphics[height=1.8cm]{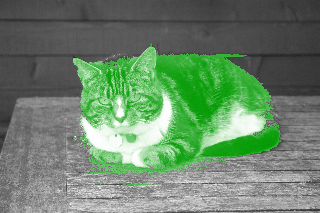}&
\includegraphics[height=1.8cm]{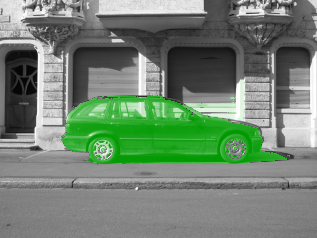}\\
\end{tabular}
\caption{Shape unary term illustrated. On the top: foreground and background likelihood shown respectively in green and red; high color values = higher likelihood. On the bottom: segmentation results.}
\label{fig:sd_unary}
\end{figure*}

\setlength\tabcolsep{4pt}
\begin{table*}[t]
\centering
\begin{tabular}{|l|c|c|c|c|c|c|c|c|c|}
\hline
 							  & \multicolumn{2}{c|}{TUD}        & \multicolumn{7}{c|}{MSRC21} 										\\ \hline
                & sideviews cars & sideviews cows & plane & cow  & car  & bike & sheep & cat  & dog \\ \hline
SharpMask       & 0.40           & 0.52 					& 0.29  & 0.71 & 0.40 & 0.19 & 0.48  & 0.61 & 0.48\\ \hline
SharpMask + MPN & 0.39           & 0.52 					& 0.29  & 0.68 & 0.37 & 0.18 & 0.44  & 0.61 & 0.45\\ \hline
BSM             & 0.39           & 0.48 					& 0.17  & 0.57 & 0.18 & 0.26 & 0.48  & 0.22 & 0.13\\ \hline
\end{tabular}
\caption{Segmentation performance ($mAP$) on TUD and \emph{things} from MSRC21}
\label{tab:mAP_TUD_MSRC21}
\end{table*}

\subsection{Segmentation}

We use the shape codebook of each codeword $C_i$ associated with a detection to compute a consolidated foreground and background hypothesis $\tilde{\upsilon}_i$. For this, we consider only the occurrences of $C_i$ that voted for the object hypothesis. Each of these occurrences is associated, through the clustering creating the shape codebook, with one of the shape descriptors $\Dsty_{i(k)}$ of that codebook. Writing $n_j$ for the index of the shape descriptor associated with occurrence $j$, we compute $\tilde{\upsilon}_i$ as
\begin{equation}
\tilde{\upsilon_i} = \sum \upsilon_{i(n_j)} w_{k,i},
\end{equation}
where the sum is computed over all occurrences of codebook $C_i$ that voted for the object hypothesis and where $\upsilon_{i(n)}$ is the strength of the foreground and background hypothesis of $\Dsty_{i(n)}$. Each occurrence is weighted by the same factor $w_{k,i}$ that was used for object detection and defined in Eq.~(\ref{eq:weight}). With this, each occurrence contributes equally to $\tilde{\upsilon}_i$ allowing all variations of the shape of an object part to be taken into consideration.

To produce pixel-wise likelihood values, we first rescale every $\tilde{\upsilon_{i}}$ to the basic training size of the descriptors\footnote{This size is initially determined in training. Empirically, a size of 21x21 pixels gave better results.} multiplied by the size of the matched feature. We then project the rescaled $\tilde{\upsilon_{i}}$ onto the matched feature location. This generates an image $I_{\tilde{\upsilon}}$ where each pixel $p$ has a value providing the likelihood of being labeled background or foreground. 

Once an object is detected and per-pixel foreground and background likelihoods are computed, we segment the object. We cast the segmentation problem as a dense CRF which we solve using the mean field approximation of Krähenbühl~\emph{et al.}~\cite{Krahenbuhl2011, Krahenbuhl2013}. The unary term $\psi_u(x_i)$ of our Gibbs energy function is defined by
\begin{equation} \psi_u(x_i) = \lambda_1 \psi_{shape}(x_i) + \lambda_2 \psi_{color}(x_i) + \lambda_3 \psi_{roi}(x_i) \end{equation}
where $\psi_{shape}$, $\psi_{color}$, and $\psi_{roi}$ are the potential provided respectively by the shape model (Sec.~\ref{sec:shape_descriptor}), kernel density estimation on the RGB channels for both foreground and background, and a region-of-interest constraint based on the object position and scale. The $\lambda$ are weighting parameters tuned at validation. We use the same pairwise term suggested by Krähenbühl~\emph{et al.}

\section{Evaluation}

We evaluate BSM performance using mean average precision (mAP~\cite{Everingham2010, Lin2014}) on the TUDarmstadt Object Dataset (TUD)~\cite{Magee2002} and MSRC21 dataset~\cite{Shotton2006}. We also provide mAP on the same datasets for a state-of-the-art deep learning approach, namely SharpMask~\cite{Pinheiro2016}, trained on MS COCO~\cite{Lin2014}. Since this dataset is several orders of magnitude larger than ours, we cannot consider SharpMask as a valuable approach for tiny datasets, even considering the possibility of fine-tuning the network. Therefore, a direct comparison between BSM and SharpMask would be unfair. However, we still find SharpMask's segmentation outputs useful as an upper bound on the achievable performance.

\subsection{Datasets}

There are two classes of objects with segmentation ground truth in TUD: \emph{sideviews-cars} and \emph{sideviews-cows}. MSRC21 contains 23 different classes; for our experiments, we have kept the \emph{thing}\footnote{The literature~\cite{Adelson2001, Everingham2010} sometimes distinguishes 2 broader categories of object classes: \emph{stuff} and \emph{things}. Stuff is associated with large areas of texture-like objects such as grass, sky, water, etc. \emph{Things} are associated with more defined objects such as cars, buildings, persons, etc.} classes to segment. We used the cleaned-up ground truth from~\cite{Malisiewicz2007}.

For both datasets, there are no predetermined sets. Considering this, we split TUD and MSRC21 in respectively 3 and 5 random folds for each class, which are combined to create different training, validation and testing sets. This way, our approach may be tested on the entire datasets; it is especially useful for the MSRC21 dataset since it has only 30 images per class. Note that, for TUD \emph{sideviews-cars} images, we were careful to keep mirrored pairs in the same fold.

\subsection{Performance}

For both methods, mAP results are shown in Table~\ref{tab:mAP_TUD_MSRC21}. Typical BSM segmentation results are also provided in Fig.~\ref{fig:sd_unary}. Since SharpMask does not produce any labeling, we funnel its segmentation masks through a MultiPath Network (MPN)~\cite{Zagoruyko2016} with a ResNet-50 feature extractor. SharpMask without MPN is evaluated solely on masks which overlap with the ground truth ($iou \geq 0.5$); other masks are discarded. Performance is evaluated only on classes which are also in COCO.

Results indicate that, for tiny training sets (as few as 18 images for MSRC21 classes), BSM performs fairly well and, for some classes, as well as a state-of-the-art approach that was trained on tens of thousands of images. Observing unary terms in Fig.~\ref{fig:sd_unary} shows that our shape descriptor creates a large margin of uncertainty around the boundaries of the object. Indeed, we consider that the boundaries location should be determined mainly by the content of the image and less by a modeled likelihood. This choice aims for less contested boundary regions; everything near a boundary would have, with this approach, weak likelihood values for the foreground and the background. In practise, this means less per-pixel false alarms and misses. Using a dense CRF model lets	the image content drive the segmentation. Our model allows us to fully exploit the strength of this method, since our shape descriptor gives low background and foreground likelihood values near object boundaries. 
Common errors are caused by large variations in appearance or shape. Shadows can create problems since the pairwise energy term models intensity variations; this can cause the resulting segmentation to incorporate the object's shadow. Another BSM limitation is that it cannot handle large scale variations.

\section{Conclusion}

We propose a shape descriptor that performs well in capturing the boundary information of an image patch. The shape descriptor can model numerous complex shapes while being robust to shape variation and small changes of pose. We use these shape descriptors to create shape models for a bag-of-feature appearance-based object detector. Our shape model identifies which regions are part of the object and those which are not, and allows the image content to drive the localization of the boundaries in the segmentation process. We show that this shape model generates very good foreground and background likelihood for detected objects which we use in a dense CRF model for object segmentation. Training this shape model necessitates few images which makes it attractive for segmentation on tiny datasets.

\clearpage
\bibliography{ICIP2017}
\bibliographystyle{hieeetr}

\end{document}